\documentclass{article} 
\usepackage{iclr2020_conference,times}

\usepackage{amsmath,amsfonts,bm}

\def\eqref#1{equation~\ref{#1}}

\def\1{\bm{1}}

\DeclareMathAlphabet{\mathsfit}{\encodingdefault}{\sfdefault}{m}{sl}
\SetMathAlphabet{\mathsfit}{bold}{\encodingdefault}{\sfdefault}{bx}{n}

\usepackage{hyperref}
\usepackage{url}
\usepackage{amsmath}
\usepackage{graphicx}
\usepackage{mwe}
\usepackage{verbatim}
\usepackage{amssymb}
\usepackage{mathtools}

\newcommand{\hide}[1]{}
\newcommand{\note}[1]{}
\renewcommand{\note}[1]{~\\\frame{\begin{minipage}[c]{\textwidth}\vspace{2pt}\center{\textcolor{red}{#1}}\vspace{2pt}\end{minipage}}\vspace{3pt}\\}

\usepackage{verbatim}
\usepackage[linesnumbered, ruled]{algorithm2e}
\SetKwRepeat{Do}{do}{while}%
\SetKwComment{Comment}{$\triangleright$\ }{}
\usepackage[normalem]{ulem}

\usepackage{wrapfig}
\usepackage{subcaption}
\usepackage{multirow}

\title{Differential Evolution for Neural Architecture Search}

\author{Noor Awad$^{1}$$^*$, Neeratyoy Mallik$^{1}$\thanks{Equal contribution}, \& Frank Hutter$^{1,2}$\\
$^1$Department of Computer Science, University of Freiburg\\
$^2$Bosch Center for Artificial Intelligence\\
Freiburg, Germany \\
\texttt{\{awad,mallik,fh\}@cs.uni-freiburg.de } \\
}

\usepackage[bottom]{footmisc}

\iclrfinalcopy 
\begin{document}

\maketitle
\begin{abstract}

Neural architecture search (NAS) methods rely on a \emph{search strategy} for deciding which architectures to evaluate next and a \emph{performance estimation strategy} for assessing their performance (e.g., using full evaluations, multi-fidelity evaluations, or the one-shot model). In this paper, we focus on the search strategy and demonstrate that the simple yet powerful evolutionary algorithm of \emph{differential evolution} (DE) yields state-of-the-art performance for NAS, comparing favourably to regularized evolution and Bayesian optimization. It yields improved and more robust results for 13 tabular NAS benchmarks based on NAS-Bench-101, NAS-Bench-1Shot1, NAS-Bench-201 and NAS-HPO bench.

\end{abstract}

\section{Introduction}


Evolutionary algorithms have a long history for neural architecture search (NAS), for combined search of architectures and weights~\citep{Stanley-nature19a, Stanley-ieee09a}, {~\citep{Mason-GECCO17a, Baioletti-Mathematics20}}, search of just the architecture~\citep{real-arXiv17a, elsken-arXiv18a, Liu-arXiv17a}, 
and multi-objective optimization of performance and resource consumption~\citep{elsken2018efficient}. Recently, regularized evolution~\citep{real-arXiv18a} has been shown to yield very robust performance on NAS benchmarks~\citep{ying-PMLR19a} and found novel neural architectures for object recognition in CIFAR-10~\citep{real-arXiv18a} and to an improved version of the transformer~\citep{evolved_transformer}.

Here, we study the use of the popular evolutionary algorithm of \emph{differential evolution} (DE, \cite{storn-springer-97a}) for NAS.
DE has previously been used to search basic neural network architectures.
For example, \cite{Mineu-IJCNN} used DE to search for layers, neurons, weights and connections for architectures using a special 
local and global neighbourhood strategy for the mutation operation, \cite{Bhuiyan-ICBIFE} introduced a simple DE algorithm without the use of a crossover operation, 
and \cite{Zhang-NC} used DE to jointly evolve architectures and weights, followed by the Levenberg-Marquardt algorithm to finetune the generated weights. Other works that use DE to evolve basic neural architectures can be found in {~\citep{Dhahri-NC12, Mason-GECCO17b}}. In this paper and different from the above, rather than developing a customized DE version for a specific task, we standardize and benchmark the use of a simple, yet effective DE, for a wide range of NAS benchmarks .  

DE has been used as one of many algorithms for a recent benchmark of joint hyperparameter optimization and NAS~\cite{Klein-ICML18}, and did not yield state-of-the-art performance there. However, that study used a simple SciPy \cite{2020SciPy-NMeth} implementation, and we demonstrate that with a better implementation and a fixed, robust hyperparameter setting, DE does indeed achieve state-of-the-art performance on a wide range of recent NAS benchmarks compared to other blackbox optimizers.

Most recent progress in NAS focuses on exploiting the one-shot model introduced by~\citet{enas}, prominently based on extensions of differentiable architecture search (DARTS~\citep{darts}). However, the one-shot model in general
\citep{sciuto} and DARTS in particular~\citep{Zela2020Understanding} feature several failure modes.
For this reason, using the terminology of \citet{elsken-jmlr19a}, we do not employ the one-shot model as a performance estimation strategy to evaluate different search strategies, but 
%
%
%
%
%
rather stick to the simpler performance estimation strategy of full evaluations. While a large-scale evaluation would normally be completely infeasible in this setting due to the high computational cost of full evaluations, this analysis is made possible by the recent availability of tabular NAS benchmarks~\citep{ying-PMLR19a}.


We first describe a canonical version of differential evolution (DE; Section \ref{DE}), then describe how to apply DE to NAS (Section \ref{DE-NAS}), and then Section \ref{sec:experiments} demonstrates that the resulting algorithm outperforms the previous best search strategies on a wide range of 13 benchmarks based on NAS-Bench-101~\citep{ying-PMLR19a} NAS-Bench-1Shot1~\citep{zela2020nasbench1shot1}, NAS-Bench-201~\citep{dong-arXiv20a}, and NAS-HPO-Bench~\citep{klein-arXiv19a}. The appendix can be found at: \href{https://ml.informatik.uni-freiburg.de/papers/20-NASICLR-DE-NAS-supplementary.pdf}{Appendix Link}.

\section{Canonical Differential Evolution}
\label{DE}
Differential Evolution (DE, \cite{storn-springer-97a}) is an evolutionary algorithm that is based on four steps (initialization, mutation, crossover and selection). We describe these below, deferring details to Appendix A. In its canonical form, DE is described for continuous optimization. 

\textbf{Initialization.} DE is a population-based meta-heuristic algorithm which consists of a population of $NP$ individuals. Each individual is considered a solution and expressed as a vector of $D$-dimensional decision variables, which are initialized uniformly at random in the search range. 

\textbf{Mutation.} A new child/offspring is produced using the mutation operation for each individual in the population by a so called mutation strategy. The classical DE uses $rand/1$ mutation, in which three random individuals/parents $X_{r_{1}}, X_{r_{2}}, X_{r_{3}}$ are chosen to generate a new vector $V_{i,g}$ as follows:
\begin{equation}
V_{i,g}= X_{r_1,g} + F \cdot (X_{r_2,g} - X_{r_3,g})
\label{Eq.2}
\end{equation}
\noindent{}where $V_{i,g}$ is the mutant vector generated for each individual $X_{i,g}$ in the population, $F$ is the scaling factor (which usually takes values within the range $[0, 1]$), and $r_1, r_2, r_3$ are the indices of different randomly selected individuals. The subscript $g$ indicates the generation index, or iteration number.

\textbf{Crossover.} After the mutation, a crossover operation is applied to each target vector $X_{i,g}$ and its corresponding mutant vector $V_{i,g}$ to generate a trial vector $U_{i,g}$. We use a simple binomial crossover, which chooses the value for each dimension $i$ from $V_{i,g}$ with probability $Cr$ and from $X_{i,g}$ otherwise.

\textbf{Selection.} After generating the trial vector $U_{i,g}$, DE computes its function value $f(U_{i,g})$, keeping $U_{i,g}$ if it performs at least as well as $X_{i,g}$ and reverting back to $X_{i,g}$ otherwise.


\section{Differential Evolution for NAS}\label{DE-NAS}
Recent NAS approaches and benchmarks parameterize cell structures of deep neural networks as directed graphs \citep{Zoph_2018_CVPR,ying-PMLR19a,zela-arXiv20a,dong-arXiv20a}. The realisation of a candidate cell structure can be seen as an assignment of operations from a set of choices or a range of values, such as the choice of operator on an edge or the choice of predecessors of a node in the directed graph. 

We found the best way of applying DE when parameters are discrete or categorical is to keep the population in a continuous space, perform canonical DE as usual as described in Section \ref{DE}, and only discretize copies of individuals to evaluate them.
If we instead dealt with a discrete population space, then the diversity of population would drop dramatically, leading to many individuals having the same parameter values; the resulting population would then have many duplicates, lowering the diversity of the difference distribution and making it hard for DE to explore effectively.  

\begin{figure*}[t]
\centering
\includegraphics[width=1.\textwidth]{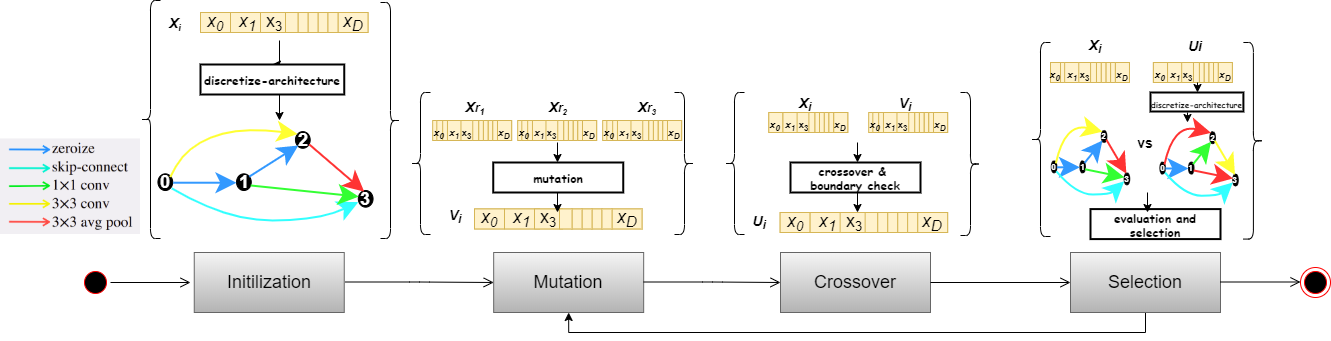} 

\caption{\label{fig:DE-Framework} DE for NAS Framework}
\end{figure*}

The modified canonical DE we used for NAS is presented in Algorithm 1 of Appendix B. Figure \ref{fig:DE-Framework} 
shows the general framework of our DE implementation. 
We scale all NAS parameters to [0, 1] to let DE work on individuals from a uniform, continuous space. 
The continuous value for $U_{i,g}$ needs to be mapped back to the original space of the NAS parameters before the function evaluation. In Algorithm 1, we use a method \textit{discretized\_architecture} to do this; this method retrieves the following values $X^i$ depending on the parameter's type:
\begin{itemize}
    \item \textit{Integer} and \textit{float} parameters: $X^i \in [a_i, b_i]$ are retrieved as: $a_i + (b_i - a_i) \cdot U_{i,g}$, where the integer parameters are additionally rounded.
    \item \textit{Ordinal} and \textit{categorical} parameters $X^i \in \{x_1, ..., x_n\}$: the range [0, 1] is divided uniformly into $n$ bins.
\end{itemize}

\begin{wrapfigure}[16]{r}{0.55\textwidth}
\vskip -0.25in
    \centering
    \includegraphics[scale=0.4]{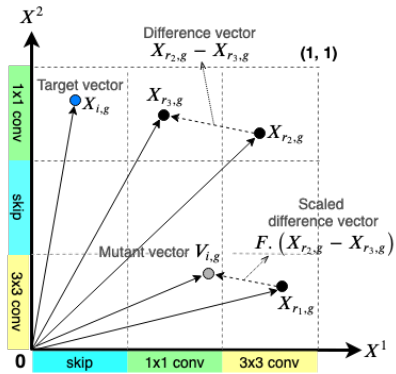}
    \caption{Illustration of DE mutation on categorical parameterization of NAS cell space}
    \label{fig:mutation}
\end{wrapfigure}

We illustrate the discretization in Figure \ref{fig:mutation}. For the categorical parameter $X^2 \in $ \{`\textit{1x1\ conv}', `\textit{skip}', `\textit{3x3\ conv}'\}, the corresponding continuous DE space maps to $[0, 1/3)$ for `\textit{1x1 conv}', $[1/3, 2/3)$ for `\textit{skip}', and $[2/3, 1]$ for `\textit{3x3 conv}'. 
As seen in Figure \ref{fig:mutation}, the \textit{difference vector} and the randomly sampled candidate individuals determine how the search space is spanned to find a \textit{mutant vector} that participates in the selection process. The resultant mutant can lie on any of the 9 grids formed in Figure \ref{fig:mutation} for the 2-dimensional case. 

One drawback for such an approach might arise in the case of a conditional parameter space. However, just like in NAS-Bench-101~\citep{ying-PMLR19a}, the function value for an \textit{invalid architecture} can simply be a maximal error of 1 (at no computational cost, even in a non-tabular benchmark). Such individuals will be guaranteed to lose in the selection process, thereby implicitly avoiding invalid architectures over time.


\section{Experiments}\label{sec:experiments}
We evaluate DE's performance on four recent NAS benchmarks: NAS-Bench-101 \citep{ying-PMLR19a}, NAS-HPO \citep{klein-arXiv19a}, NAS-Bench-1shot1 \citep{zela-arXiv20a} and NAS-Bench-201 \citep{dong-arXiv20a}. We compare against several baseline algorithms, namely Random Search (RS) \citep{bergstra-jmlr12a}, BOHB \citep{falkner-icml18a}, Tree Parzen Estimator (TPE) \citep{bergstra-nips11a}, Hyperband (HB) \citep{li-jmlr18a} and regularized evolution (RE) \citep{real-arXiv18a}. Appendix C has more details about the used algorithms and their hyperparameter settings. For DE, we set scaling factor $F$ and crossover rate $Cr$ to 0.5 over all the generations. For the population size $NP$, we tested several values (provided in Appendix F and chose 20 for our experiments. We consider RE as the \textit{primary} baseline algorithm (run until $10M$s) since it belongs to the same family of algorithms as DE and has been shown to perform robustly many times before. We provide a comparison of the robustness between RE and DE in Appendix E.
For each algorithm, we performed 500 independent runs and report the mean performance of the immediate validation regret \citep{ying-PMLR19a}.
Throughout, we evaluate algorithms in the anytime setting, showing performance of the best found configuration over time as suggested by~\citet{ying-PMLR19a} and \citet{lindauer_best_practices}.
In all our plots, the x-axis shows \textit{estimated wall-clock time}, as the cumulative time taken for training each of the architectures found as returned by the NAS benchmarks. 
Due to the space limitation, we show the test regret plots for NAS-101 and NAS-1shot1, and also discuss the experiments for NAS-Bench-201 and NAS-HPO in Appendix D.1, D.2, D.3, D.4, respectively. We compare our implementation with the popular SciPy-DE code \cite{2020SciPy-NMeth} in Appendix G. Our code for DE and for reproducing our experiments
is publicly available at \url{https://github.com/automl/DE-NAS}.

\subsection{NAS-Bench-101}
\begin{figure}
\centering
\begin{subfigure}{.32\textwidth}
  \centering
  \includegraphics[width=\textwidth]{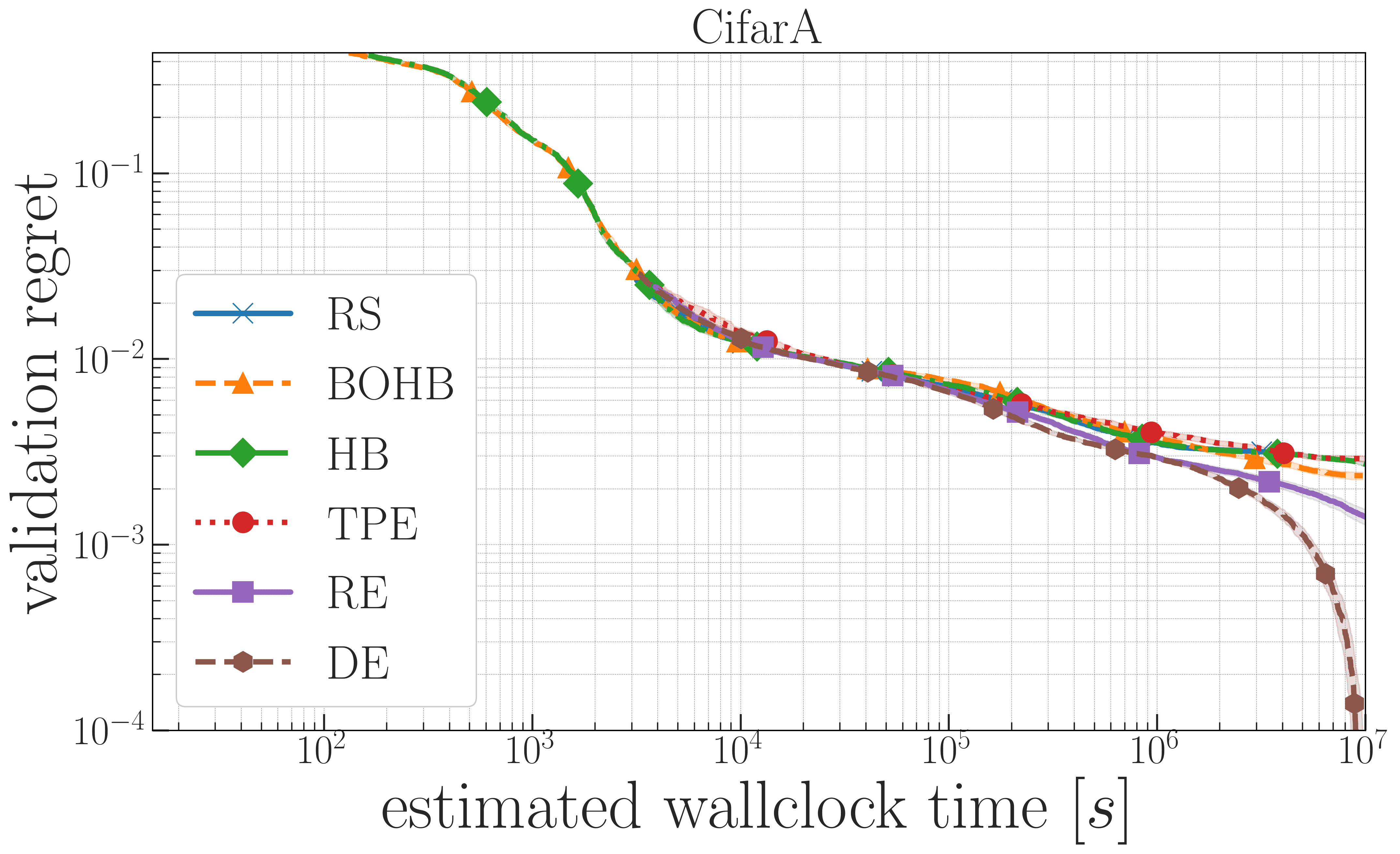}
  \caption{CifarA}
\end{subfigure}%
\begin{subfigure}{.32\textwidth}
  \centering
  \includegraphics[width=\textwidth]{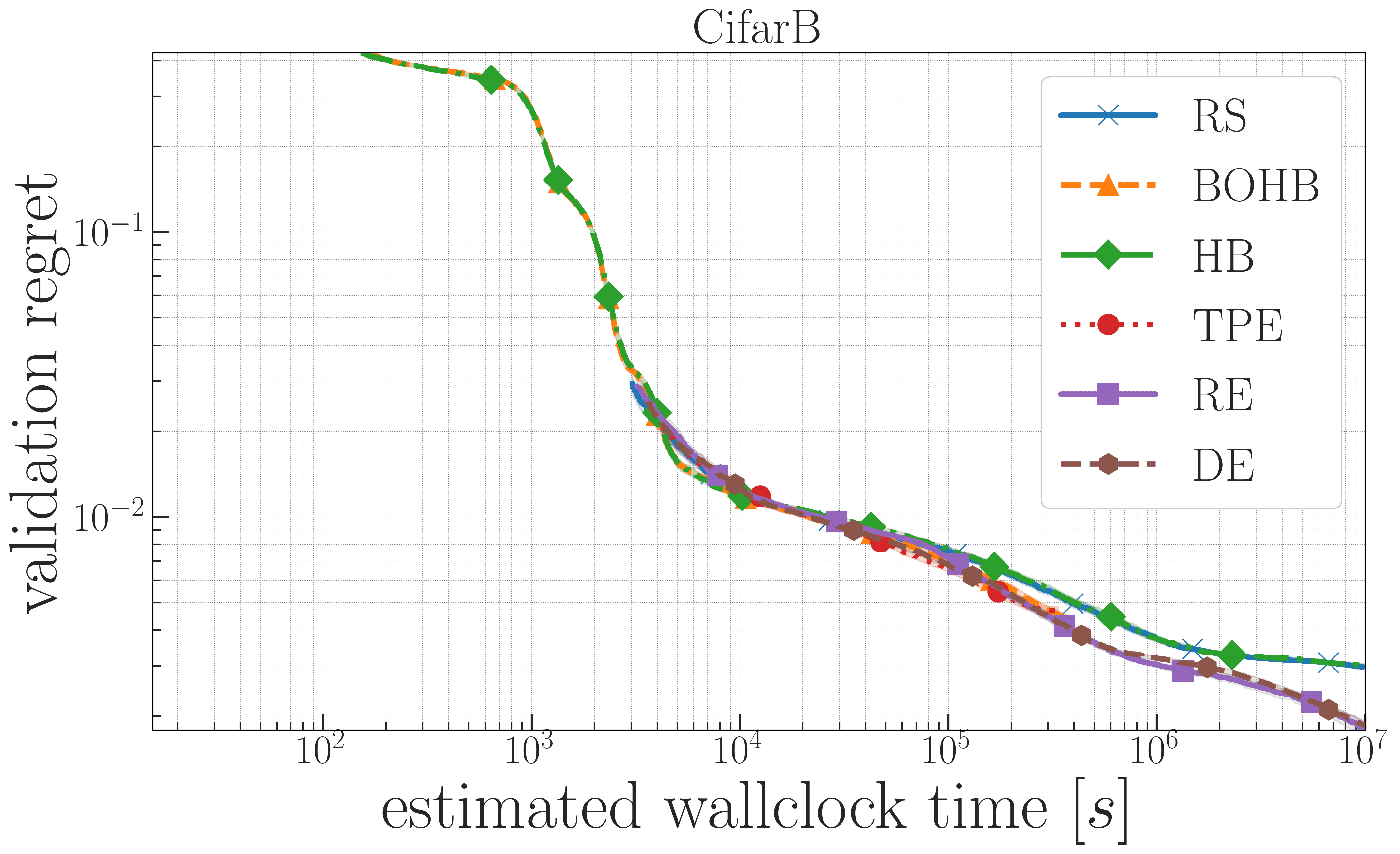}
  \caption{CifarB}
\end{subfigure}
\begin{subfigure}{.32\textwidth}
  \centering
  \includegraphics[width=\textwidth]{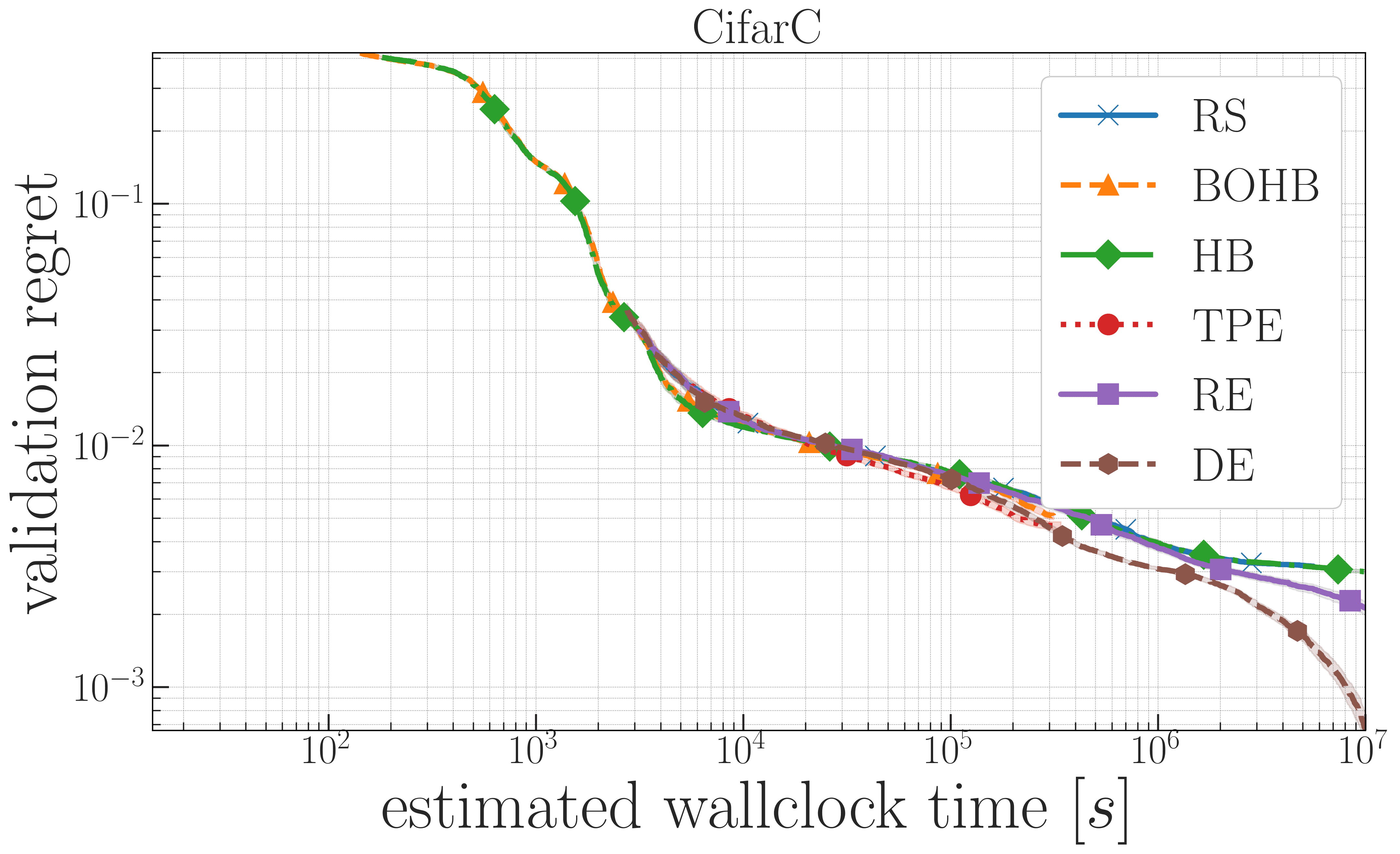}
  \caption{CifarC}
\end{subfigure}
\caption{A comparison of the mean validation regret performance of 500 independent runs as a function of estimated training time for NAS-Bench-101 on CifarA, CifarB and CifarC.}
\label{plot:NAS-101}
\end{figure}

In this experiment we investigated DE's performance on the cell search space of 423k unique cell architectures of a convolutional neural network for CIFAR-10 defined by NAS-Bench-101~\citep{ying-PMLR19a}.
We study three different search spaces: 
\textit{CifarA} contains the main search space discussed by \citet{ying-PMLR19a}, and \textit{CifarB} and \textit{CifarC} are variants of the same space with alternative encodings (treating the edge parameters as categorical parameters with 21 choices and continuous $\in$ [0, 1], respectively).
Figure \ref{plot:NAS-101} presents a comparison of the performance of compared algorithms showing the mean validation regret of 500 independent runs as a function of the estimated training time. We show our results for test regret in Appendix D.1. 
HB and BOHB are multi-fidelity optimization algorithms which evaluate at fewer epochs while other algorithms evaluate only at $E_{max}$. However, NAS-Bench-101 features a low rank correlation between the performance obtained with different budgets~\citep{ying-PMLR19a}, and thus these algorithms do not perform better than the other algorithms that only use the maximum number of epoch.
The other algorithms (RS, TPE, and RE) follow the same behaviour at the beginning of the search for all 3 encodings of the search space, and in the end the evolutionary algorithms RE and DE clearly yield the best performance. DE shows much better final performance for \textit{CifarA} and \textit{CifarC} and competitive performance with RE for \textit{CifarB}. 
It appears that DE is able to exploit high-dimensional spaces well and handle mixed-types better. This may be attributed to NAS-Bench-101's locality property \citep{ying-PMLR19a} along with DE's search method, since a DE population with individuals from a good region will be able to exploit further and get near the global optimum.

\subsection{NAS-Bench-1Shot1}
\begin{figure}
\centering
\begin{subfigure}{.32\textwidth}
  \centering
  \includegraphics[width=\textwidth]{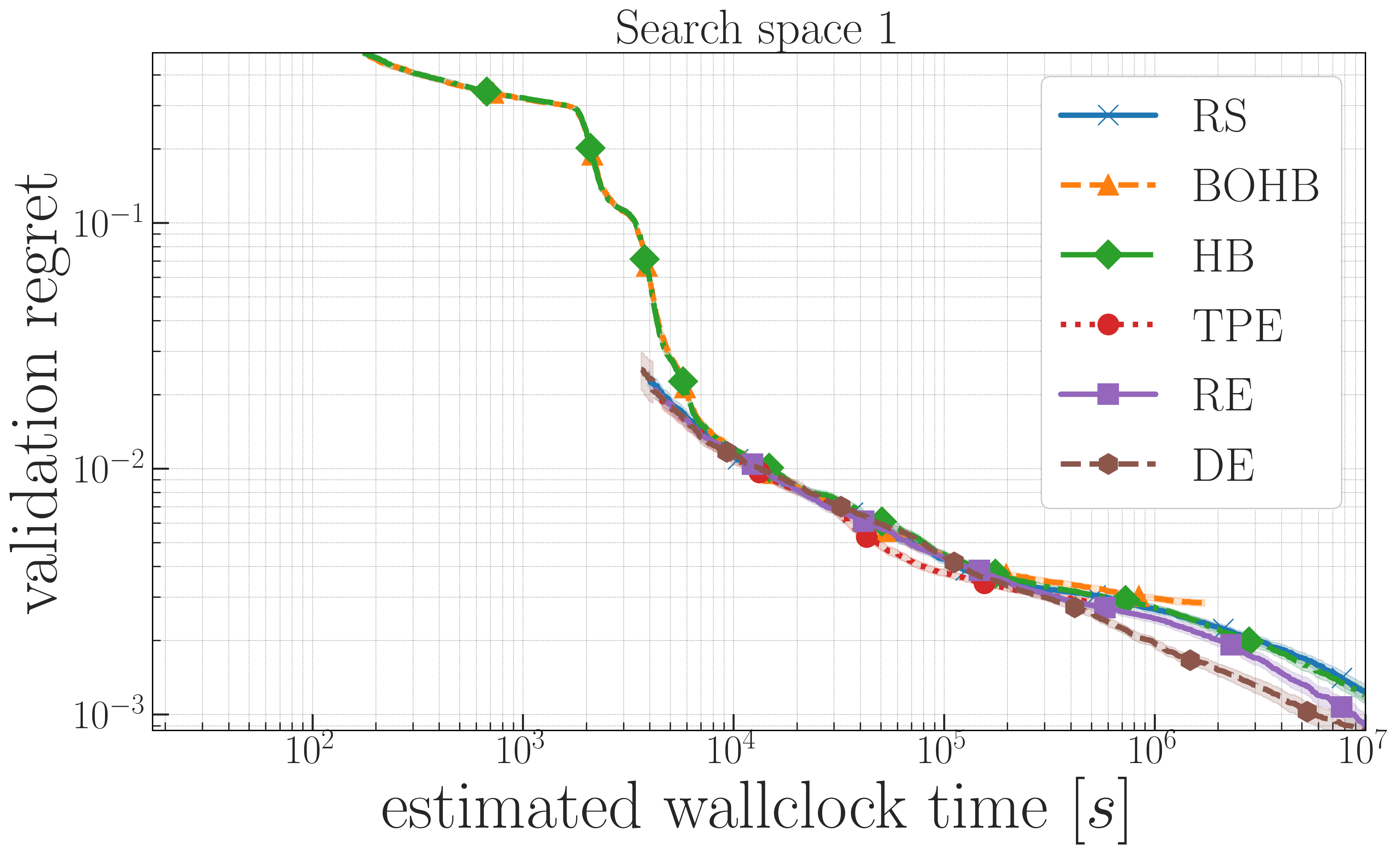}
  \caption{Search Space 1}
\end{subfigure}%
\begin{subfigure}{.32\textwidth}
  \centering
  \includegraphics[width=\textwidth]{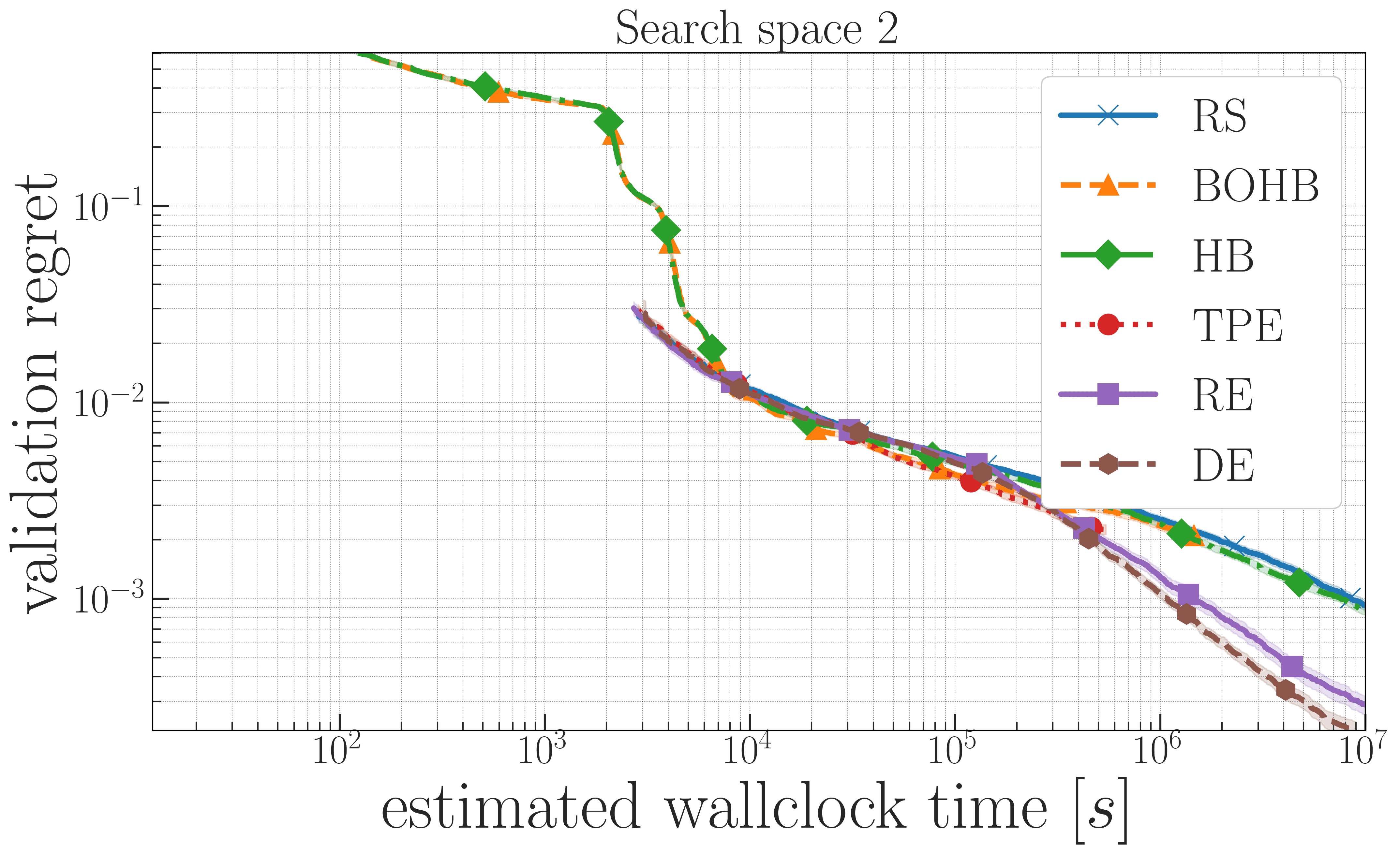}
  \caption{Search Space 2}
\end{subfigure}
\begin{subfigure}{.32\textwidth}
  \centering
  \includegraphics[width=\textwidth]{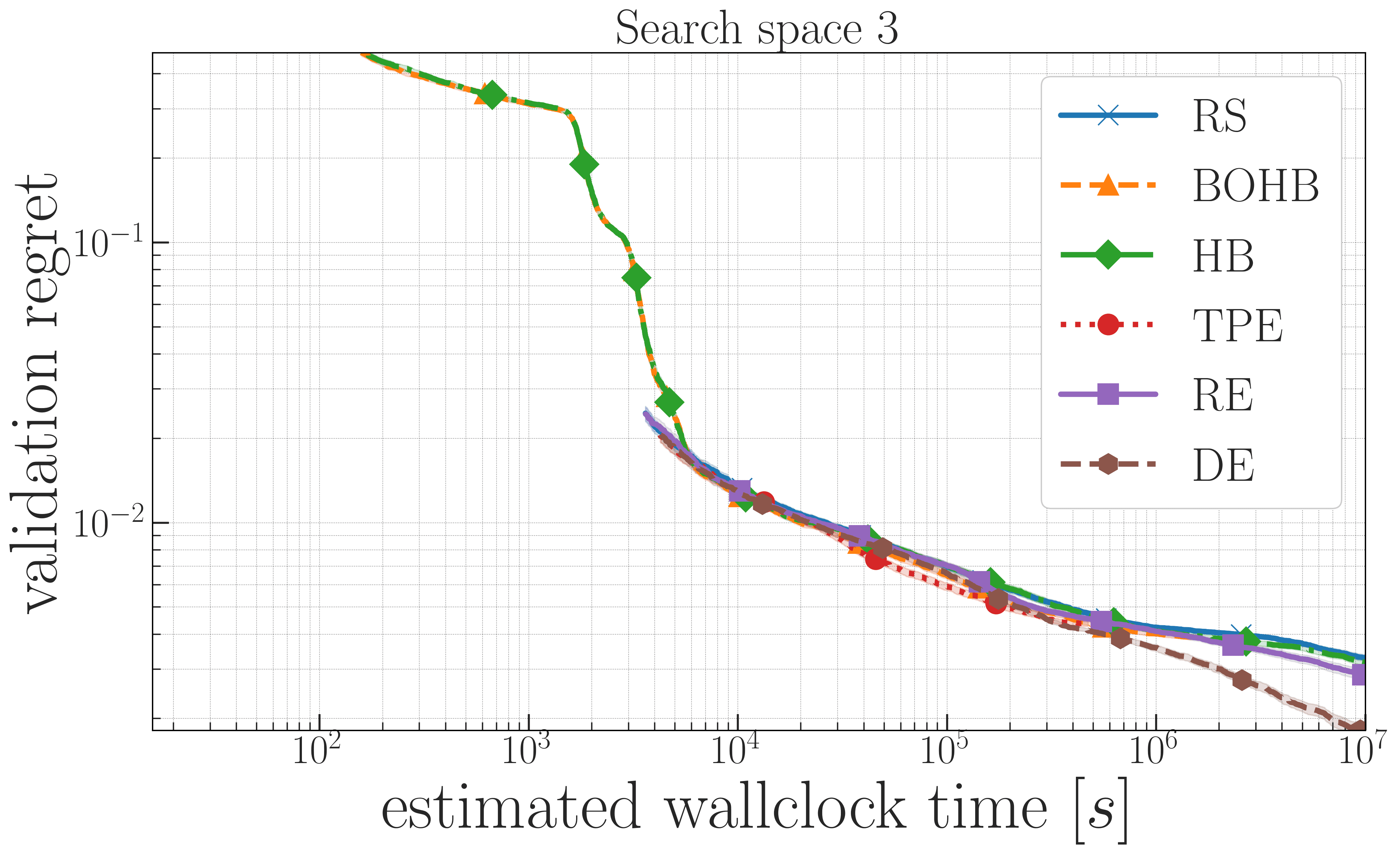}
  \caption{Search Space 3}
\end{subfigure}
\caption{A comparison of the mean validation regret performance of 500 independent runs as a function of estimated training time for NAS-1Shot1 on the three different search spaces.}
\label{plot:NAS-1shot1}
\end{figure}

NAS-Bench-1Shot1~\citep{zela2020nasbench1shot1} was created from the search space of NAS-Bench-101 by keeping the network-level topology intact and modifying the cell-level topology to allow the application of modern weight sharing algorithms for three search spaces with
$6,240$ (search space 1), $29,160$ (search space 2), and $363,648$ (search space 3) architectures. Figure \ref{plot:NAS-1shot1} shows our results for the mean performance on validation regret while we present a comparison on test regret in Appendix D.2.
For search space 1, all the algorithms achieve nearly the same error at the beginning of the search, then DE converges faster until other algorithms catch up. For search space 2, RE and DE converge fastest. 
For search space 3, the most complex (high-dimensional) and largest (10x more architectures than space 2, and 100x more than space 1), DE clearly outperforms all other algorithms and converges fastest. 


\section{Conclusion}
We demonstrated that Differential Evolution can be utilised as an alternative search strategy for the growing field of NAS. 
We also demonstrated DE's ability to handle mixed data types and high-dimensional spaces.
DE may thus be a good candidate for NAS in very large spaces that may help discover new, yet unknown, architectural design patterns. 
Since DE naturally lends itself well to parallelization, future work includes providing a parallel implementation. We are also interested in combining DE with different performance estimation strategies, such as multi-fidelity methods and the one-shot model. Our reference implementation of the code is available at \url{https://github.com/automl/DE-NAS}.

\section*{Acknowledgments}
The authors acknowledge funding by the Robert Bosch GmbH for financial support.

\clearpage
\bibliography{shortstrings,local,lib,shortproc}
\bibliographystyle{iclr2020_conference}


\end{document}